\documentclass[lettersize,journal]{IEEEtran}
\usepackage{amsmath,amsfonts}
\usepackage{algorithmic}
\usepackage{algorithm}
\usepackage{array}
\usepackage[caption=false,font=normalsize,labelfont=sf,textfont=sf]{subfig}
\usepackage{textcomp}
\usepackage{stfloats}
\usepackage{url}
\usepackage{verbatim}
\usepackage{graphicx}
\usepackage{cite}

\usepackage{booktabs}

\begin{document}

\title{Time Scale Network: A Shallow Neural Network For Time Series Data}

\author{ Trevor Meyer, Camden Shultz, Najim Dehak, Laureano Moro-Vel\'azquez,  Pedro Irazoqui
        % <-this % stops a space
\thanks{This work has been submitted to the IEEE for possible publication. Copyright may be transferred without notice, after which this version may no longer be accessible.}
\thanks{This work was funded by NIH NS119390.}% <-this % stops a space
\thanks{All authors are with the Department of Electrical and Computer Engineering at Johns Hopkins University}% <-this % stops a space
\thanks{Manuscript received 28, June, 2023; revised X, X, 202X.}}

\markboth{Journal of Selected Topics in Signal Processing (UNDER REVIEW)}%
{Shell \MakeLowercase{\textit{et al.}}: A Sample Article Using IEEEtran.cls for IEEE Journals}

\IEEEpubid{\copyright~2023 IEEE}
%\IEEEpubid{0000--0000/00\$00.00~\copyright~2023 IEEE}

\maketitle

\begin{abstract}
% The abstract must be between 150-250 words.
Time series data is often composed of information at multiple time scales, particularly in biomedical data. While numerous deep learning strategies exist to capture this information, many make networks larger, require more data, are more demanding to compute, and are difficult to interpret. This limits their usefulness in real-world applications facing even modest computational or data constraints and can further complicate their translation into practice. We present a minimal, computationally efficient Time Scale Network combining the translation and dilation sequence used in discrete wavelet transforms with traditional convolutional neural networks and back-propagation. The network simultaneously learns features at many time scales for sequence classification with significantly reduced parameters and operations. We demonstrate advantages in Atrial Dysfunction detection including: superior accuracy-per-parameter and accuracy-per-operation, fast training and inference speeds, and visualization and interpretation of learned patterns in atrial dysfunction detection on ECG signals. We also demonstrate impressive performance in seizure prediction using EEG signals. Our network isolated a few time scales that could be strategically selected to achieve 90.9\% accuracy using only 1,133 active parameters and consistently converged on pulsatile waveform shapes. This method does not rest on any constraints or assumptions regarding signal content and could be leveraged in any area of time series analysis dealing with signals containing features at many time scales.
\end{abstract}


\section{Introduction}
\label{sec:intro}

\IEEEPARstart{A}{pplied} signal processing systems introduce many constraints that must be considered in algorithm development. Particularly, when working in biomedical data systems, we encounter many limitations not seen in other fields like computer vision (CV) and natural language processing (NLP). Many state-of-the-art CV and NLP architectures and techniques rely on continually increasing network architecture size and complexity based on expected signal structure and content, require large datasets and heavy computational resources to train and deploy, and suffer from reduced overall intepretability. Meanwhile, biomedical applications are unable to use the same approaches as they frequently face severely limited data availability which can be of low quality~\cite{wang_deep_2019}, ideally incorporate processing on local devices both to enable rapid response during medical events and to address security in handling sensitive health information~\cite{basatneh_health_2018}, and benefit greatly from interpretable outputs that provide meaningful feedback to patients or clinicians using the algorithm to evaluate and manage disease~\cite{ching_opportunities_2018}.

Due to these limitations, accuracy cannot be the only metric considered for the successful translation of algorithms in the biomedical space. For practical application in real-world settings, it is essential to take into account the available data resources for network training and the computational capabilities of devices beyond the GPU-focused servers commonly used by many deep learning (DL) researchers and design algorithms for those applications. This includes scenarios where there may be insufficient data to train large networks or lack of expert knowledge about the embedded patterns to effectively hand-tune specialized techniques. This also includes applications that utilize CPU-powered computers and low-power edge devices, like cellphones and wearable/implantable medical devices, which face strict memory and computational maximums that signal processing and machine learning (ML) systems should balance. While some edge applications have relied heavily on Application Specific Integrated Circuits (ASICs), these are not realistic solutions for everyone as these chips are incredibly time consuming and expensive to create and modify. 

Furthermore, explainable and interpretable algorithms will aid the translation to real-world applications. More explainable algorithms lead to a better understanding of potential failure modes of said algorithms and resulting decisions, enabling application specific mitigation strategies that can increase confidence and efficacy. Additionally, they can enable knowledge discovery and research directions in evolving disease research~\cite{ching_opportunities_2018}. None of the aforementioned attributes are adequately characterized by just a reported percent accuracy. Our goal in this publication is to balance all these considerations to better enable the translation of DL algorithms directly into clinical use rather than focusing on improving accuracy by a few percentage points.

 \IEEEpubidadjcol
One promising approach for enabling DL algorithms in this context is to adapt and refine existing validated structures to be more efficient and effective by considering and combining both classical signal processing frameworks and newer DL techniques. Here, we present a novel architecture that we call the Time Scale (TiSc) Network. It draws inspiration from wavelets by combining the translation and dilation sequence used in discrete wavelet transforms (DWTs) with traditional convolutional neural networks (CNNs) and back-propagation to create a shallow neural network that maximizes accuracy-per-parameter and accuracy-per-operation. We also are able to use network saliency techniques to produce "visual explanations" of network decisions that can lead to insights in network utilization and signal content/structure. We initially demonstrate these characteristics on interpretable electrocardiogram (ECG) signals to detect atrial dysfunction where we assess performance in comparison to other shallow networks and show key advantages, including interpretability and fast training and inference. Then, we demonstrate strong performance on less interpretable electroencephalogram (EEG) signals to predict seizure onset, and can identify dominant feature scales that can be used to repeatably extract waveforms with interpretable features and further minimize network parameter counts below 1,200 parameters without sacrificing accuracy. Model architectures and implementations are freely available.~\footnote{Available upon request. Contact tmeyer16@jhu.edu}

\section{Time-Frequency Analysis}
\label{sec:timefrequencyanalysis}

Time series signals are well described by the concepts of frequency and superposition, particularly in the medical field where biological signals are composed of many physiological components that oscillate or recur at some independent rate and combine to form a measured signal. Therefore, time-frequency analysis techniques are quite effective and are incorporated into many widely used processing systems. These include: spectrograms, which compute frequency descriptors over a limited time window; modified spectrograms like Mel-Frequency spectrums, which average information within certain frequency bands; and wavelet decompositions, which optimize resolution-in-frequency with resolution-in-time in a generalized, invertible transform.

While Fourier-based techniques are widely used, they have some well-documented weaknesses such as sensitivity to small deformations or time-warping at high frequencies, and struggle to characterize long-duration patterns due to reduced resolution at low frequencies~\cite{anden_deep_2014}. While increasing the duration analyzed by a Fourier Transform can increase the resolution at low frequencies, this will sacrifice the temporal resolution of the entire transformation and dilute the specificity of short-term features as more time and variation is accumulated. This will also result in oversampling of the higher frequency information, creating many unnecessary or redundant coefficients. 

Wavelet transforms better manage these phenomena and can create feature-rich embeddings of artifacts of many different time scales. DWTs are generalized invertible transforms that decompose a signal by discretely translating and dilating a “mother wavelet” to extract time-localized frequency content. High-frequency content is more finely sampled in time by a shorter mother wavelet and low-frequency content is more finely sampled in frequency by an exponentially lengthening mother wavelet. This elegant time-frequency lattice has proven especially powerful when applied to real-world signals and is used in applications ranging from compression (JPEG 2000) to signal analysis. For example, when applied to ECG signals, a well-designed wavelet transform can simultaneously capture characteristics of impulse-like QRS complexes, slower P and T waves, still slower heart rate, and even very low frequency patterns such as respiration rate in a single transform~\cite{pal_detection_2010}.

There has been a vast exploration of the best wavelet designs to use in medical contexts which satisfy orthogonality, symmetry, frequency isolation, and other constraints that are central to wavelet theory. However, we find these constraints limiting when used to fit a classification objective. For example, all wavelets ensure perfect reconstruction and, therefore, embed all variation in any signal. To efficiently extract information most optimized classification, it is often not necessary to capture all variation. In fact, capturing all variation will likely add noise to the embedding as much of the variation will be irrelevant for accurate differentiation of each class. To better allow feature extraction specifically optimized for a classification objective, we harness the power of DL and back-propagation to design custom wavelet shapes at many time scales without numerical constraints.

\section{Prior Literature}
\label{sec:priorliterature}

The integration of wavelet-inspired techniques and ML can be seen as early as the 1990s with adaptive wavelet techniques~\cite{szu_neural_1992,pati_analysis_1993}, wavelet networks ~\cite{zhang_wavelet_1992,zhang_wavelet_1995,zhang_using_1997}, and wavelet-based graphs~\cite{crovella_graph_2003,jansen_multiscale_2009}. Other more classical CNN-style architectures aim to achieve the similar multi-scale feature extraction~\cite{cui_multi-scale_2016,wang_multilevel_2018,chen_deep_2020,liu_adaptive_2020}, including dilated convolution~\cite{yu_multi-scale_2016}, invariant scattering~\cite{bruna_invariant_2012}, temporal convolution networks~\cite{lea_temporal_2016,zhang_new_2018}. Each approach is plagued with various constraints, such as reliance on a pre-defined wavelet basis, restriction to an orthonormal basis, incorporation of other mathematical constraints, down/sub-sampling schemes that do not incorporate full-resolution input, processing only a few hand-selected window sizes, otherwise constrained feature representation, or are highly application specific.

Contrary to these approaches, the goal of our network is to find a non-parametric transform that utilizes a particular translation/dilation operation without being numerically constrained to a solution space while avoiding any specific data-based assumptions. More specifically, we do not constrain our learned weights to be invertible or orthogonal or otherwise satisfy any previously mentioned constraints, and we do not require linear or time invariant solutions, and we do not assume any Fourier-derived conclusions including using sampling theory to justify sub-sampling schemes, relying entirely on the robustness of back-propagation to converge on a solution. This keeps our approach from being task-specific, so even though this solution was designed specifically for biomedical signal content, it still maintains relevance to any signal processing application that considers signal components at different time scales.

In addition to wavelet-based methods, there are many memory-based structures that aim to capture similar short and long-term features, including recurrent neural network components like long short-term memory units~\cite{hochreiter_long_1997}, Legendre memory units~\cite{voelker_legendre_2019}, and other recurrent structures. The integration of memory units is another way to solve many of the problems discussed here. However, these architectures can be complicated by requiring strictly sequential data samples during training, and take significantly more operations to compute, leading to long training and inference times. Also, these components are not at all interpretable.

Lin et al. presents a similar network to ours called the "Feature Pyramid Network" that they use to detect objects at various size scales in CV~\cite{lin_feature_2017}. While it is successful in CV and has also been used to analyze spatial features of time series signals~\cite{hou_deep_2022}, this technique has not been adapted and optimized for to extract time series patterns across many time scales.

\begin{figure*}[t]
\begin{minipage}[b]{1.0\linewidth}
  \centering
  \label{fig:tisc}
  \centerline{\includegraphics[width=17.5cm]{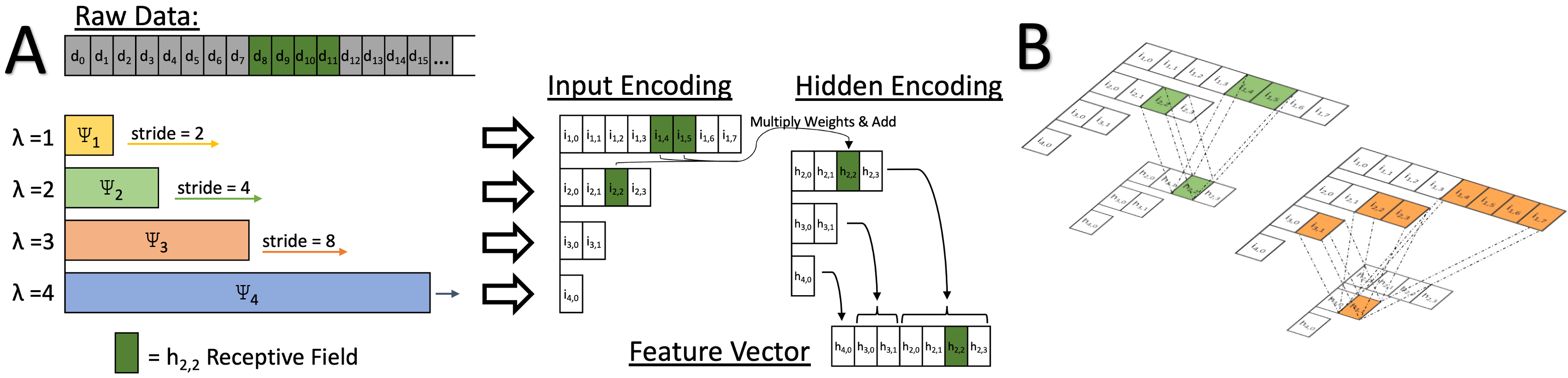}}
%   \vspace{2.0cm}
%   \centerline{(a) Result 1}\medskip
    \caption{(A) Illustration of TiSc input layer (d to i) and TiSc hidden layer (i to h) operation. In the TiSc input layer, learned waveforms $\Psi$ of exponentially increasing length slide across the raw data with a stride equal to their length. In the TiSc hidden layer, activations are combined according to their receptive field in the raw data, assuring only activations contained in a particular time scale are combined. The receptive field of $h_{1,2}$ is colored to demonstrate this. (B) Further visualization of the TiSc hidden layer activation combination scheme for two scales $\lambda=2$ (green) and $\lambda=3$ (orange). An activation in a deeper layer (below) is created by combining several activations from the previous layer which span the associated time scale.}
\end{minipage}
\end{figure*}

\section{Time Scale Network}
\label{sec:timescalenetwork}

In this study, we introduce the Time Scale Network, or TiSc Net, to simultaneously and efficiently incorporate features at many different time scales. TiSc Net implements strided convolution, following a similar computational structure to the DWT. Waveform kernels are learned via back-propogation without any numerical constraints, where each scale is independent from the others, allowing for optimized feature extraction specific to the task at hand. The input layer is summarized in Equation 1, where waveforms of exponentially increasing length are translated across the data $x$ with a stride equal to the window size. The output of each windowed multiplication is saved as a new activation. The TiSc input layer activations at time scale $\lambda$ and offset $i$ are calculated as

\begin{equation}
    \label{eq:tisc}
    % \sum_{\lambda \in \Lambda} \sum_{n=0}^{L/2^\lambda} 
    \begin{aligned}
    g( \Psi_\lambda \cdot (x[n-2^\lambda i] \cdot w[2^\lambda]) )  \\
    \forall{\lambda \in \Lambda} \text{   } , \text{   } \forall{i \in [\text{}0 , L/2^\lambda\text{})}
    \end{aligned}
\end{equation} 
where $\Lambda$ is limited to integers in the range $[1,log_2 (L)]$, $i$ are positive integers, $L$ is the length of the input data, $\Psi_\lambda$ is a learned waveform of length $2^\lambda$ specific to scale $
lambda$, $g$ is a non-linear activation funcion, $\cdot$ is the dot product, and $w$ is defined as
\begin{equation}
    w[m] = \begin{cases}
            1 & 0 \leq n < m \\
            0 & otherwise
    \end{cases}
\end{equation}

This translation-dilation pattern creates receptive fields equivalent to those of individual coefficients of a DWT. A notable difference between our method and  wavelet transforms is that each scale waveform is applied to $x$ in parallel rather than sequentially. Also, higher-scale wavelet shapes are not subsampled, so they incorporate full-resolution inputs. This is an intentional choice to assure long-term patterns are processed in full detail, rather than limiting to features which remain after subsampling.

This translation-dilation sequence is continued within the embedded space where TiSc Hidden Layers combine activations according to the same time-scaled receptive field, assuring consideration of strictly time-limited features. An illustration of this is shown in Figure~1. The resulting equation is the same as Equation \ref{eq:tisc}, but $x$ represents the embedded space and $w[m]$ selects activations at all $\lambda$ whose receptive fields are entirely contained in $0\leq n<m$ in the raw data. These hidden layers can be repeated or stacked to create deeper or wider networks. The last TiSc hidden layer does not have a nonlinearity before it is passed to additional classification layers, typically a dense connection to a one-hot output.

Multi-channel data can be incorporated into this architecture by interleaving values from each channel into a single vector and increasing $\Lambda$ values accordingly to maintain the same time-resolution for each $\lambda$. This elegantly maintains the above structure while allowing each $\Psi_\lambda$ to incorporate time-aligned data from all channels in a single operation.

It is worth highlighting the efficiencies of this algorithm. For each scale $\lambda$, the entire inner product for all $i$ can be calculated by reshaping the input data matrix to rows of length $2^\lambda$ and applying matrix-vector multiplication with the weight vector, a computation with significant hardware acceleration in GPUs, CPUs, and even micro-processors. The entire computational sequence for a single input layer is most efficiently implemented as shown in Algorithm~\ref{alg:compute}, where each time scale output is fully calculated by a single matrix multiplication. Further, despite the non-square structure, we avoid the need for sparse arrays or other data management overheads when storing weights and activations by leveraging the fact that these grow with the same dimensionality as a binary (or n-ary) tree and inverted binary tree, respectively. We follow standard binary tree memory organization where all values are stored along one dimension and indexed such that moving between scales is possible through multiplication by 2 and moving within each scale is possible by adding the offset.

\begin{algorithm}[t]
    \centering
    \caption{Time Scale Computation}
    \label{alg:compute}
    \begin{algorithmic}[1]
        \STATE $s \leftarrow \lambda\_min$
        \STATE $w \leftarrow 2^s$
        \STATE $i \leftarrow length(data) / w$
        % \STATE $kernel \leftarrow length(data) / w$
        \WHILE{ $scale <= \lambda\_max $}
        \STATE reshape data $d$ to have rows of length $w$
        \STATE $k \leftarrow weights[w-1 : w*2-1]$
        \STATE $r \leftarrow k.matmul(d) + bias[s]$
        \STATE $out[i-1 : i*2-1] \leftarrow r$
        \STATE $w \leftarrow w * 2$
        \STATE $i \leftarrow w / 2$
        \STATE $s \leftarrow s + 1$
        \ENDWHILE
        \RETURN $out$
    \end{algorithmic}
\end{algorithm}

The computational complexity of this algorithm, or number of required operations as a function of input size, is bounded between $O(n)$ and $O(n lgn)$, and its constants are easily expanded or reduced by adjusting $\Lambda$. This represents a significant advantage over convolution operations in spectrograms or CNNs, which are bounded between $O(n lgn)$ and $O(n^2)$ depending on dimension choices and use of fast implementations. The number of stored parameters and activations are also easily changed by adjusting $\Lambda$, with integer increments altering the number of parameters and activations by a factor of 2 each increment. 

When a network shows strong performance, it is sometimes desirable to visualize the learned feature kernels and consider the importance of each feature. Due to its versatility, we chose the gradient-weighted class activation mapping (GradCAM)~\cite{selvaraju_grad-cam_2017} method to extract what time scales were most important for classification and when relevant features appear in the raw data. GradCAM uses the back-propagated gradient magnitudes to determine which features most significantly contributed to a particular output, and is not specific to any architecture. These gradients can visualized within a single example or further be summed over an entire dataset to summarize the importance of a particular time scale in general.

\section{Results}
\label{sec:results}

To demonstrate the advantages of TiSc Net, we choose two application areas as case studies: Atrial Dysfunction (AD) detection using two-channel ECG signals and Seizure Prediction (SP) using 16 channel EEG signals. Case studies allow us to consider the usage of our network in the context of a real application, enabling in-depth discussion of specific details, nuances, and insights regarding the data rather than relying on generalized metrics, such as accuracy, which fail to adequately represent many of the trade-offs we are interested in. This approach also better reflects the considerations involved in the translation to clinical practice, and better demonstrates why TiSc Net has more utility in these scenarios than other networks.

\subsection{Atrial Dysfunction Detection}

AD is a type of cardiac dysfunction in which the atrial chamber of the heart fails to contract properly. One classic dataset containing examples of AD through ECG data is the MIT-BIH Arrhythmia Database that we used with TiSc Net to classify examples as either healthy or AD~\cite{moody_mit-bih_1992}.

We extracted 2 channels of ECG data (lead orientation was variable as expected in clinical practice) randomly from 47 subjects as non-overlapping segments of length 2.845 seconds (2048 samples) labeled as either healthy, AFIB (atrial fibrillation), AB (atrial bigeminy), or AFL (atrial flutter). There were more healthy segments than AD segments, so we randomly sub-selected from the healthy segments to achieve a balanced dataset of 10,770 examples for each class. All networks were trained to convergence with the rmsprop optimizer, regularization, and 5\% dropout using 10-fold cross-validation on 70\% of the samples. The remaining 30\% of samples were used to test the converged networks, and only these accuracies are reported in following sections. Validation and testing accuracies rarely differed by more than 1\% unless overall performance was poor.

\subsubsection{Time Scale Network}

We first trained a network consisting of one TiSc input layer with $\Lambda=[2,11]$, then one TiSc hidden layer with $\Lambda=[3,11]$, and a dense connection to one-hot outputs designating the segment as healthy or AD. We conducted a hyperparameter search, sweeping the range of $\Lambda$, the presence of hidden layers, and the number of parallel TiSc channels. We found many successful networks which were robust across a range of selected hyperparameters, with results plotted in the first row of Figure~2A. For example, we found one network that balanced memory and computational considerations used a $\Lambda=[6,9]$ with three independent channels, achieving 81\% accuracy with 6,947 parameters and 25,329 multiply-accumulate operations (MACs).

\begin{figure}[t]
\begin{minipage}[b]{1.0\linewidth}
  \centering
  \label{fig:accuracy}
  \centerline{\includegraphics[width=8.5cm]{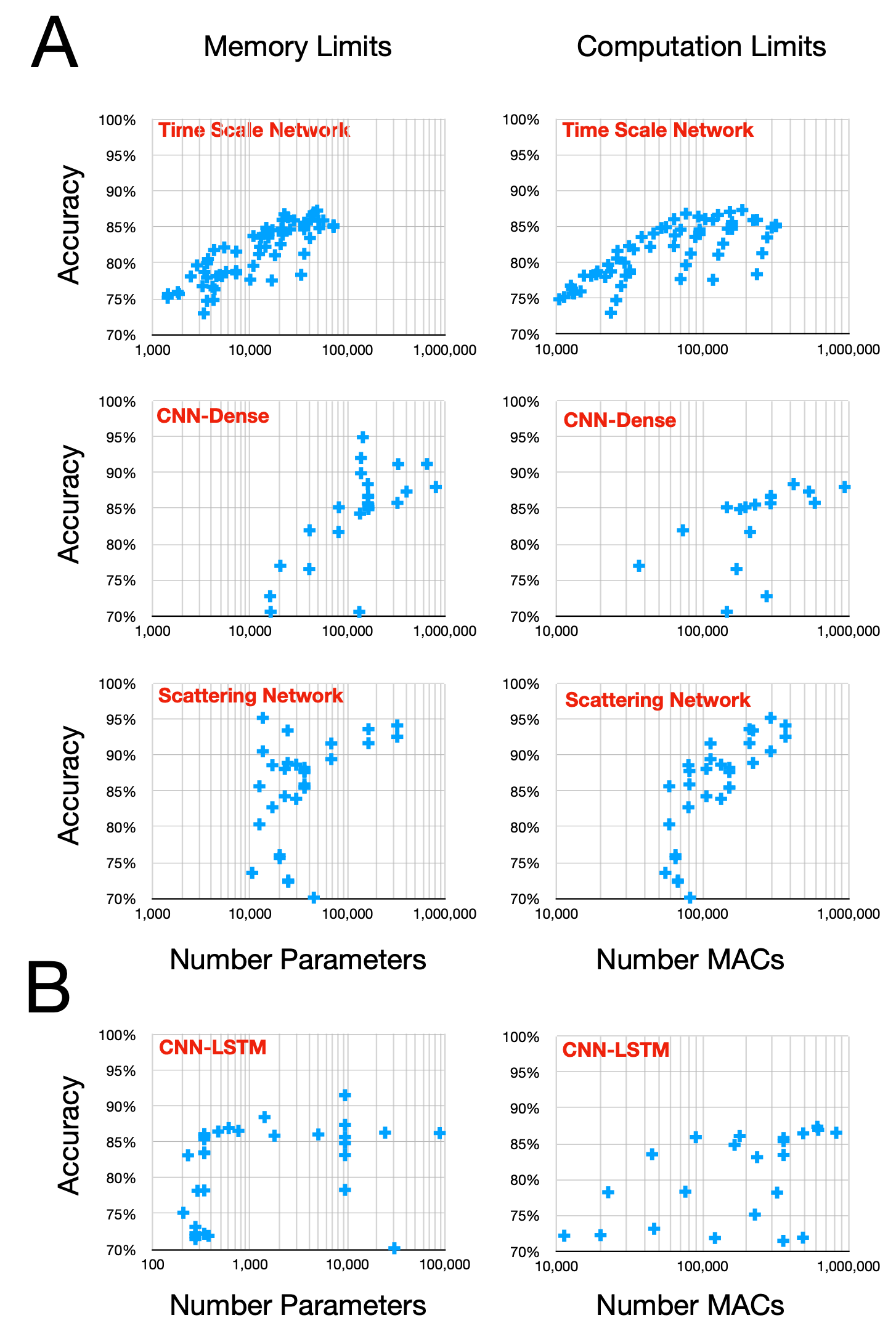}}
%   \vspace{2.0cm}
%   \centerline{(a) Result 1}\medskip
    \caption{Accuracy-per-parameter and accuracy-per-operation (MACs) for a range of optimized network architectures. The best networks approach the top left corner, maximizing accuracy while minimizing parameters and/or MACs. The x-axes differ between A \& B.}
\end{minipage}
\end{figure}

\subsubsection{Other Architectures}

While there are many successful networks presented on this dataset, none are as efficient as ours in terms of number of parameters or operations required. Instead, existing literature typically presents large networks that do not optimize for parameter and computational considerations. Therefore, to better compare the limits of the capabilities of other network types on this dataset, we implemented a wide range of shallow CNNs and wavelet-based networks with explicit optimization of parameter and computation constraints for a more thorough comparison of each technique within our objectives.

% \paragraph{Convolutional Neural Network}

We constructed many CNNs consisting of convolutional, pooling, and densely connected layers producing a one-hot output (CNN-Dense). While a single CNN layer densely connected to a one-hot output performed poorly (below 75\% accuracy), adding an additional convolutional, pooling, or dense layer improved accuracy to a relevant level. For these network structures, we searched hyperparameters, including the convolutional or pooling window length, stride, number of channels, and size of the hidden dense layer. While CNN-based architectures demonstrated the highest accuracies, their accuracy quickly decreased as we reduced the number of parameters and MACs, as shown in Figure~2. Notably, we found the most effective way to reduce operations was to increase stride, a strategy that is already optimally implemented in TiSc Net.

% \paragraph{Wavelet Scattering}

The best non-recurrent architecture we tested was the wavelet scattering network ("Deep Scattering Network")~\cite{anden_deep_2014}. We utilized a wavelet scattering transform densely connected to a single hidden layer followed by a one-hot output. We searched hyperparameters, including the wavelets per octave (Q), maximum log-scale (J), and size of the hidden layer. We again see a significant decrease in accuracy as the number of parameters are reduced, as shown in Figure~2.

% \paragraph{Recurrent Networks}

We implemented CNNs with LSTM cells (CNN-LSTM) in place of dense layers, as LSTM architectures boast low parameter counts with high accuracy. We searched along the same hyperparameters as in CNN-Dense with the addition of searching many hidden state sizes. While we saw excellent performance with very few parameters, CNN-LSTM networks typically required several orders of magnitude more operations to compute, with many network MACs beyond the upper limit of the plot shown in Figure~2B. While accuracy-per-parameter benchmarks of CNN-LSTM networks did beat TiSc Net, they suffered greatly in accuracy-per-operation. We again find the best way to reduce the number of operations is by increasing the stride, but CNN-LSTM networks resulted in reduced accuracy compared to TiSc Net when choosing comparable stride attributes.

An exhaustive list of all the architectures we tested and their performances can be found in the Supplementary Information.

\begin{table}[b]
\begin{center}
  \label{table:timeit}
  \caption{ Computation time of DL architectures \\  MacBook Pro M1 Max - CPU only - averaged across 100 trials.}
  % \begin{tabular}{ l | c | c }
  %   % \hline
  %   Architecture & Inference Time & Epoch Time \\
  %   \hline
  %   TiSc      & 0.492 ms & \bf{1.149 s} \\
  %   CNN-Dense & \bf{0.088 ms} & 2.823 s \\
  %   Scatter   & 0.879 ms & 3.118 s \\
  %   CNN-LSTM  & 3.351 ms & 3.007 s \\
    
  \begin{tabular}{ l | r | r | c | c }
    % \hline
    Architecture & Params & MACs & Inference Time & Epoch Time \\
    \hline
    TiSc      &  6,947 & 25,329 & 0.492 ms & \bf{1.149 s} \\
    CNN-Dense & 40,776 & 73,262 &  \bf{0.088 ms} & 2.823 s \\
    Scatter   & 12,448 & 59,028 & 0.879 ms & 3.118 s \\
    CNN-LSTM  &    338 & 45,064 & 3.351 ms & 3.007 s \\
    \\
    % \hline
  \end{tabular}
  
  \caption{After hyperparameter sweeps, the networks from each architecture with the minimum number of MACs that also demonstrated at least 80\% accuracy were chosen for benchmarking. CNN-Dense used a 2x8 kernel (stride 2) with a 20 unit dense layer; CNN-LSTM used a 2x8 kernel (dilation 8, stride 8) and a Bi-LSTM with 8 hidden units; Scatter used a transform J=4, Q=1, order=2, and a 10 unit dense layer; and TiSc used 3 channels with $\Lambda=[1,5]$, all with one-hot output.}
\end{center}
\end{table}

\subsubsection{Computation Speed}

We selected networks from each architecture demonstrating at least 80\% accuracy with the minimum number of MACs. In Table 1, we report the single sample inference time and single epoch training time on a MacBook Pro M1 Max (CPU only). We find the TiSc net trains the fastest, performing training epochs in 41\% of the time of the next best network, and infers the fastest, predicting in 15\% of the time of a comparable LSTM network despite its reduced parameter count, indicating computational bottlenecks both in training and at inference when using LSTM cells. This demonstrates the significant impact of TiSc Net which not only reduces the overall parameters and MACs, but does so in an intentional way with efficient implementations that maintain a relevant level of accuracy; allowing TiSc Net to greatly reduce the training and inference time endured by researchers and patients.

\subsubsection{Interpretability}

Furthermore, we can interpret specific conclusions about our signal of interest from TiSc net, an increasingly desired attribute in medicine. We use the network saliency tool GradCAM to visualize what parts of the input were most important to specific classification outputs from TiSc Net~\cite{selvaraju_grad-cam_2017}. As shown in Figure~3, it is clear that when comparing feature saliency of an example of a healthy ECG with that of one during AD, the network utilizes time-localized features at different time scales. Further, we can observe increased feature saliency from times between the QRS complexes when making an AD classification, with the most prevalent time occurring shortly \emph{after} the QRS impulse. Although AD is caused by abnormalities in atrial firing \emph{before} the QRS impulse, it is known that AD also results in changes in ventricular filling which is visible after the QRS impulse~\cite{brignole_av_2021}. These post-firing characteristics interestingly appear more salient in AD detection in some examples.

\begin{figure}[t]
\begin{minipage}[b]{1.0\linewidth}
  \centering
  \label{fig:gradcam}
  \centerline{\includegraphics[width=8.5cm]{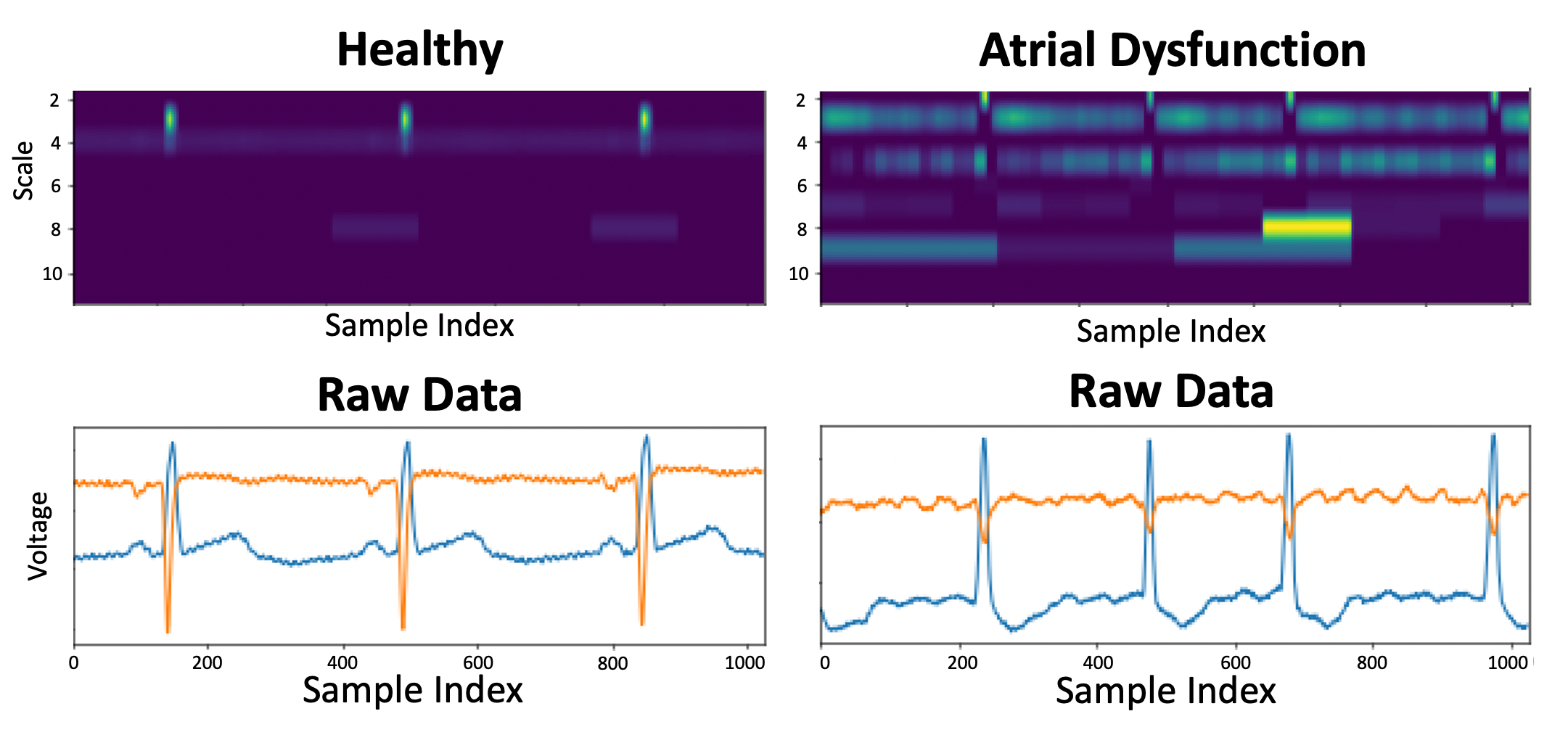}}
%   \vspace{2.0cm}
%   \centerline{(a) Result 1}\medskip
    \caption{(LEFT TOP) GradCAM output from a healthy ECG segment. The network is attending to the small-scale, short-term features of the QRS complex. (LEFT BOTTOM) Time-aligned raw data containing two leads of ECG. (RIGHT TOP) GradCAM output from an AD ECG segment. The network is attending to larger-scale features between the QRS complex. (RIGHT BOTTOM) Time-aligned raw data containing two leads of AD ECG. Sample Rate is 360 Hz. }
\end{minipage}
\end{figure}

\subsection{Seizure Prediction}
\label{sec:seizure}

Another biomedical classification objective is SP using EEG signals. Seizures are characterized by the presence of abnormal, rhythmic, and synchronized firing of the brain. Experts in the field hypothesize that you may be able to predict seizure onset before it occurs by detecting "unstable" or "high risk" brain states in the minutes to hours prior~\cite{stirling_seizure_2020}. One classic dataset containing EEG signals from patients before, during, and after seizure is the CHB-MIT dataset, which we used with TiSc Net to classify examples as either baseline or pre-ictal (before seizure)~\cite{shoeb_application_2009}.

It is important to note in this application that seizure characteristics are known to be highly specific to each patient, as in they do not follow any generalized pattern across the patient population. This means subjects should be analyzed independently, limiting the amount of data available to train a single network. There is also limited knowledge of how to interpret EEG. While the physics of EEG signals are well-know, the dynamics of how brain activity culminates to produce the local field potentials captured by surface electrodes is not well understood or dynamically modeled by domain experts~\cite{jefferys_advances_2010}. EEG analysis in practice is mostly limited to either power in specific frequency bands ranging from 1 Hz to 30 Hz or the presence of "burst" activity. This complicates the design of ML approaches, since relevant features beyond these are not known and often change between patients or even within a single patient, and presents a great opportunity for personalized DL strategies, specifically multi-scale approaches such as TiSc Net.

We extracted 16 channels of EEG data (channels specified in Supplementary Information) that was either 4 hours before/after the start/end of a seizure (baseline) or within 1 hour of seizure onset (pre-ictal) as non-overlapping segments of length 4 seconds (16,384 samples). We had more baseline examples than pre-ictal, so we randomly select an equivalent amount of baseline examples as there are pre-ictal examples for that subject. Sample counts were different for each subject, but averaged 3,460 examples for each baseline and pre-ictal class (with a range from 439 to 8,515). All networks were trained to convergence with the rmsprop optimizer, regularization, and 5\% dropout using 10-fold cross-validation on 70\% of the samples. The remaining 30\% of samples were used to test the converged networks and only these accuracies are reported in the following sections. Validation and testing accuracies rarely
differed by more than 1\% unless overall performance was poor. We only report average performance across 21 subjects in this dataset. We excluded subjects CHB08 and CHB12 as due to the timing of these subjects' seizures they contained no baseline segments according to our labeling scheme and CHB24 due to irregularities in channel T8-P8.

\begin{figure}[b]
\begin{minipage}[b]{1.0\linewidth}
  \centering
  \label{fig:gradcam}
  \centerline{\includegraphics[width=8.5cm]{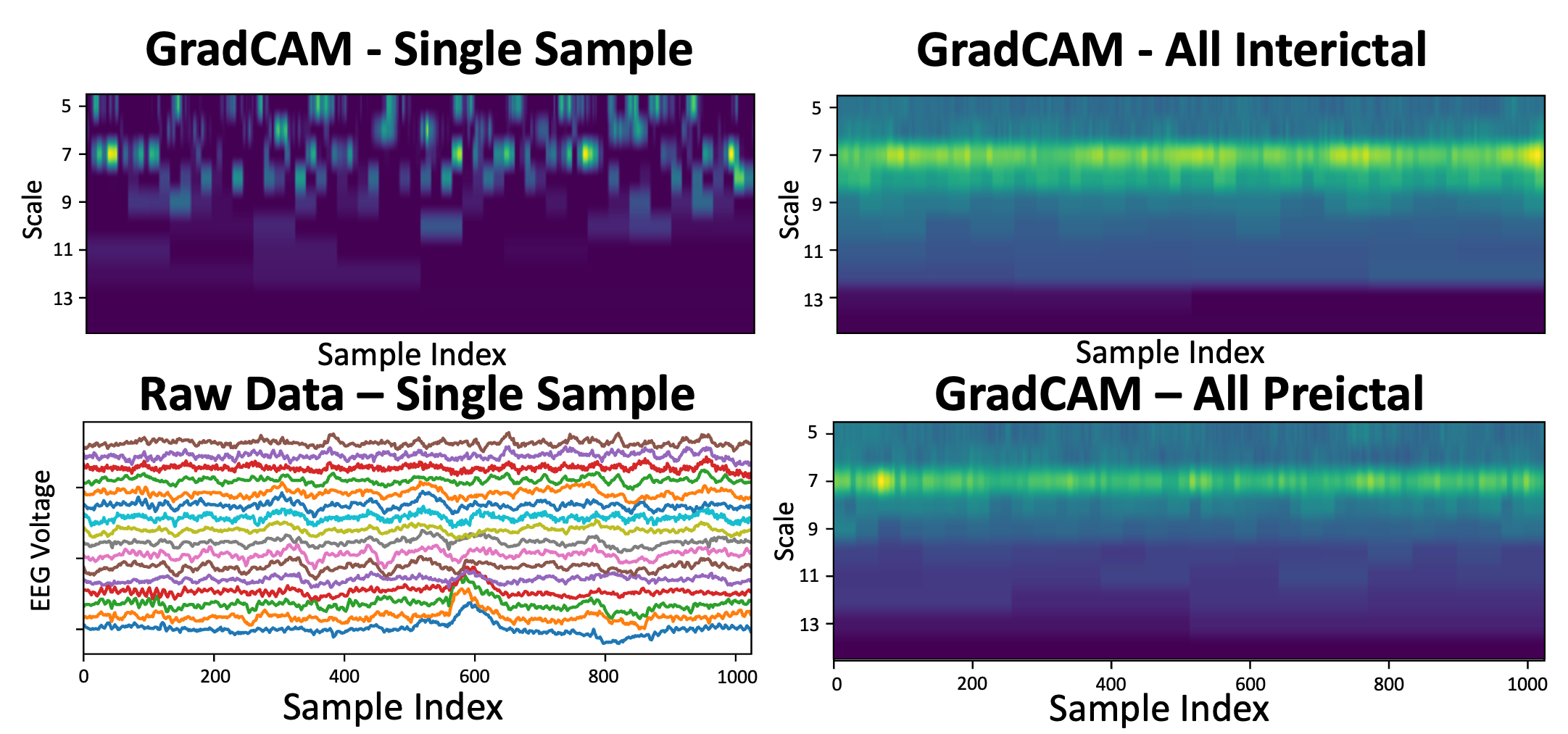}}
%   \vspace{2.0cm}
%   \centerline{(a) Result 1}\medskip
    \caption{(LEFT TOP) GradCAM output from a single EEG segment. The relevant features that are impacting the network output are much less interpretable with EEG data. (LEFT BOTTOM) Time-aligned raw data containing 16 channels of EEG data. (RIGHT TOP) Cumulative GradCAM output from every segment for subject CHB03, which had near-average accuracy and near-average number of segments available to train. It is clear that across all segments, the network rarely utilizes excessively small or large time scales, focusing primarily on scales of length 256, 512, and 1024 (Note: length of 256 = 16 samples x 16 channels)}
\end{minipage}
\end{figure}

\subsubsection{Time Scale Network}
We first trained a network consisting of one TiSc input layer with $\Lambda=[5,14]$, then one TiSc hidden layer with $\Lambda=[6,14]$, and a dense connection to a one-hot output designating the segment as baseline or pre-ictal. We searched hyperparameters including a range of $\Lambda$, the presence of hidden layers, and the number of independent channels. We found the network that best optimized accuracy contained three TiSc channels each containing an input layer with $\Lambda=[5,10]$, a hidden layer with $\Lambda=[6,10]$, all densely connected to a one-hot output which achieved a 92.3\% accuracy with 10,274 parameters and 209,904 MACs.

One interesting pattern identified during hyperparameter tuning of these networks which was not true for the AD networks is smaller networks with less parameter counts tended to have higher accuracy. In fact, when observing the GradCAM output summarizing feature saliency across the entire dataset, it was clear that, even when all the available time scales were included, the network was heavily utilizing time scales of length 256 and 512 (corresponding to window lengths of 62.5 ms and 125 ms, respectively, due to the interleaving of 16 channels sampled at 256 Hz). Therefore, we reduced a TiSc Net to include only these scales, minimizing the network to only 1,422 parameters and 33,021 MACs. Amazingly, we still achieved 90.9\% accuracy in an under 200 epochs. The parameter and operation counts could be further minimized by breaking from our data organization scheme, which follows that of binary tree structures and therefore requires placeholder values to occupy lower indices even if those coefficients are unused. If we superficially remove the unused coefficients, we find that we reach this accuracy with only 1,133 active parameters and 32,979 non-zero multiply accumulate operations. When considering CHB06, a subject picked for it's average accuracy of 90.4\% and maximum dataset size of 8,515 samples in each class, our TiSc network converges quickly to reach 85\% accuracy in an average of 20 epochs with a batch size of 256. On a Macbook Pro M1 Max (CPU only), each subject could be trained to 85\% accuracy in under 20 s, and reaches maximum accuracy in under 5 minutes.

Given the short duration of these features, we were concerned about the lack of time-invariace in TiSc Net, which might cause the network to miss the expression of any unaligned features. We implemented a network with equivalent kernel sizes but using standard convolution operations (stride=1,2,3...). When reducing the maximum stride of each scale down to 1, we saw little-to-no performance increase, yet an explosion in number of parameters and operations required. This implies that precisely time-invariant processing may not be critical for this dataset or in general, and that time-invariance constraints may enforce redundant oversampling and over-processing of signal attributes, which can be greatly reduced without significant impact on performance.

We also attempted to apply pre-training techniques to increase performance, including pre-training the TiSc layers on other subjects ahead of time and implementing a TiSc encoder-decoder architecture to learn compressed latent representations before considering classification objectives. Each of these strategies resulted in equivalent or slightly worse performance when compared to a randomly initialized network, so when considering SP these strategies are discarded.

\subsubsection{Other Architectures}
We were able to sufficiently replicate some published networks reporting success with this dataset, seeing equivalent reported accuracy on some subjects, however we observed that these algorithms were much more sensitive to architecture changes than ones we tested with the MIT-BIH dataset. Because of this we saw limited success when trying to generate new networks with comparable parameter counts, operation counts, and overall accuracy, meaning a comprehensive analysis like we previously showed would yield little insight. We can, however, compare our network performance to other networks as they are presented in literature as a best-case comparison. As shown in Table~3, we are able to achieve competitive accuracy with a reduction in parameter counts by an order of magnitude or more.

\subsubsection{Interpretability}
In addition the the GradCAM observations mentioned in previous sections, which were used to drastically reduce network size, we also observed patterns in the waveforms $\Psi$ learned by our network. Within the same subject, un-seeded, randomly initialized networks trained on slightly different folds of training data often converged on similar waveforms, as seen in Figure~5 which overlays the learned weights of 10 different training folds. This occurred in multiple subjects and was robust to a wide ranges of hyperparameter selections. This implies that for seizure prediction, in a significant proportion of subjects, there is a short ~10ms impulse occurring around 16 Hz which is highly relevant to predicting seizure onset. We reserve presentation of a deeper analysis and consideration of potential physiological mechanisms behind this phenomenon to a future publication that will include more biological and physiological background.

\begin{table}[t]
\begin{center}
  \label{table:compareNetworks}
  \caption{Comparison of DL networks tested on the CHB-MIT dataset. Accuracies marked with "*" are sensitivities, not overall accuracies.}
  \begin{tabular}{ l | l | l | l }
    Study  & Network     & Acc & Number Params \\
    \hline
    Khan 2018~\cite{khan_focal_2018}             & Wavelet+CNN+Dense  & *87 &  150,693  +DWT \\  %  *87.8\% 
    Truong 2018~\cite{truong_convolutional_2018} & STFT+CNN+Dense     & *81 &  164,653  +FFT \\  %  *81.2\% 
    Daoud 2019~\cite{daoud_efficient_2019}       & Dense              & 83  & 8,870,291 \\       %  83.63\% 
                                                 & CNN+Dense          & 94  &   520,477 \\       %  94.10\% 
                                                 & AE+Bi-LSTM+CS      & 99  &    18,345 \\       %  99.66\% 
    Tsiouris 2018~\cite{tsiouris_long_2018}      & Preproc+LSTM+Dense & 99  & 4,863+FFT+DWT \\ %  99.63\% 
    {\bf This Work}                              & TiSc+Dense         & 92 &          10,274 \\ %  88.90\% 
                                                 & TiSc+Dense         & 91  &      {\bf 1,422} \\ %  88.90\% 
    
    % Study & Accuracy & Number Params \\
    % \hline
    % Truong et. al. 2018~\cite{truong_convolutional_2018} & *81.2\% &   164,653  +FFT \\
    % Khan et. al. 2018~\cite{khan_focal_2018}             & *87.8\% &   150,693  +DWT \\
    % Daoud et. al. 2019~\cite{daoud_efficient_2019}       & 83.63\% & 8,870,291 \\
    %                                                      & 94.10\% &   520,477 \\
    %                                                      & 99.66\% &    18,345 \\
    % Tsiouris et. al. 2018~\cite{tsiouris_long_2018}      & 99.63\% &     4,863  +FFT+DWT \\
    % {\bf This Work}                                 & {\bf 88.90\%} & {\bf 960} \\
  \end{tabular}
\end{center}
\end{table}

\begin{figure}[b]
\begin{minipage}[t]{1.0\linewidth}
  \centering
  \label{fig:waveform}
  \centerline{\includegraphics[width=8.5cm]{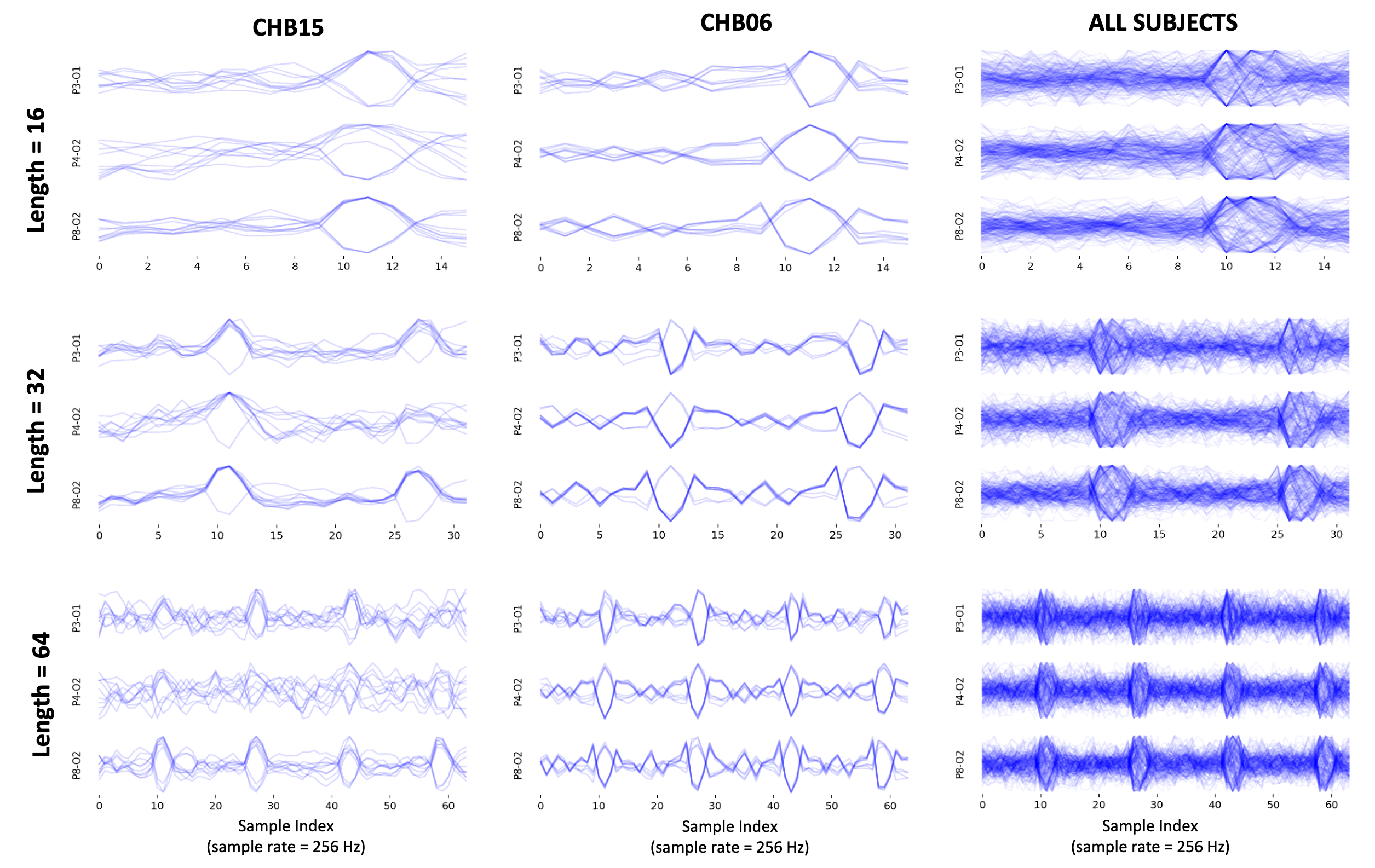}}
    \caption{Example learned waveforms from TiSc input layers. Each blue trace is learned from  un-seeded, randomly-initialized weights from one fold out of a 10-fold cross-validation scheme. We show multiple waveforms from different folds to emphasize the consistency of this learned feature, both within subjects and surprisingly between multiple subjects. This 16 Hz spiking pattern is not extracted in every subject, but is present in a large proportion. Note that as shown in CHB06, some waveforms are inverted but still characteristically similar.}
\end{minipage}
\end{figure}

\section{Discussion}
\label{sec:discussion}

\subsection{Time Scale Considerations}

We believe one primary benefit of this architecture is that it provides efficient, simultaneous consideration of long-term and short-term features in a single layer. This architecture enables full-scale receptive fields immediately after the first layer, keeping long-term features from having to be processed through many sequential layers of short-duration kernels as in CNNs, altered by smaller scales or subsampling schemes as in wavelet transforms, or obscured in memory parameters as in LSTMs. Additionally, the hidden layers combine features of different scales which occur together, allowing for consideration of these independent features in the greater context of the rest of the signal. It also incorporates long-term features without significant computational burden. In fact, this architecture significantly reduces the computational burden compared to CNN and LSTM methods. This architecture also spans short and long-term features using exponentially increasing windows, an efficient approach that more densely samples lower frequencies where many relevant biomedical patterns are expressed.

This network can rely on very few parameters, which can reduce over-fitting, particularly when there is limited data from a limited number of subjects, as is often the case when dealing with medical data or when developing personalized algorithms for patient-specific solutions. It simultaneously enables translation of these algorithms to many applications facing limitations in available memory and computation power.

This network does not rely on any assumptions or constraints regarding feature content, and it is not limited to a specific application. It does not require extensive knowledge about input signals and boasts potential for interpretable conclusions after the fact, a huge advantage for the fields of science and medicine who are hoping to use DL in research settings where, unlike in CV and NLP, the relevant features may not be known or well-understood. The simple and interpretable aspects promote transparency and trust in these algorithms by non-experts, which is becoming a growing problem in regulatory bodies who are hesitant to translate DL approaches into clinical settings. 

\subsection{Towards the Edge}

To show a key application space benefiting from our approach, we reference one demonstration of DL on a general purpose microcontroller implementing LeNet-5 using 61,706 8-bit parameters and  executing 416,520 MACs, resulting in 0.26 frames per second or a classification every 4 seconds~\cite{unlu_efficient_2020}. While an encouraging proof-of-concept, this presentation implements a very simple CV network; requires parameter quantization; uses memory-optimizations like in-place computation; does not have inference speed ideal for real-time applications; and has poor battery life. Our best network only needs 2.3\% of the parameters. Therefore, there is no need for quantization which reduces accuracy, or laborious implementations of in-memory computation. It also only requires 8\% of the MACs, which would drastically improve inference time, incredibly valuable in medical contexts. TiSc Net is advantageous in that parameters and operations are minimized from the start. Rather than applying last-minute fixes to high-performance networks to squeeze them into memory, we address these limitations from the beginning through strategic architecture designs.

\subsection{Limitations}

This network is limited to single or multi-channel time series data and will perform the best when operating with signals containing either a variety of short and long duration characteristics or unknown features that need to be identified and extracted. This is specifically applicable in any scenario where macro-level trends result from or may be obscured by micro-level actions, such as biology, economics, sociological analysis, internet activity, environmental trends, and other dynamics-based systems. Also, given its exponential structure, it works best when input lengths and number of channels can be reduced/expanded to a power of two. If for example only three channels of data exist, a fourth channel of all zeros can be used to satisfy dimensional requirements, but this results in wasted computation and parameter storage in our predefined structure.

This network is likely less useful in systems where extensive information is known about the signal characteristics, such that algorithms can be specifically engineered to target such features like in NLP. Also, objectives involving sparse event detection are likely better suited by different approaches like convolution, as they are more explicitly time-invariant.

\subsection{Perfect Performance}

It is worth discussing briefly that, in our opinion, many disciplines of science and medicine would benefit much more from a simple and interpretable network with 80\% accuracy than a complicated black-box algorithm with 99\% accuracy. Not only are uninterpretable algorithms not useful for researchers trying to use DL uncover new findings, but in translational settings, many are understandably hesitant to trust a tool that is not transparent about its function or failure modes, yet claims to be perfect. This scenario is causing friction and distrust, inhibiting and delaying the incorporation of such algorithms into regulated applications~\cite{zhu_2021_2022}. Moreover, there are many situations in medicine where classification labels are imperfect and/or debated by experts, such as whether brain activity is truly "abnormal" and should be labeled as a seizure, further undercutting the objective of obtaining 100\% accuracy. In this case, it is likely only possible by overfitting to potentially arbitrary or debatable labels. We believe our structure supports a kind of simplicity and transparency that is more likely to be relevant and accepted by non-experts even if more sophisticated proposals do claim higher accuracies.

\section{Conclusion}

Applied signal processing systems require us to consider more than just accuracy to balance constraints from real-world hardware. TiSc Net demonstrates maximum accuracy-per-compute and accuracy-per-parameter when compared to modern architectures and has shown both AD detection and SP faster and more efficiently than any other architecture, potentially saving millions of lives. Finally, we show interpretable characteristics, which are increasingly in demand in science and medicine, where harnessing DL and its sophisticated signal processing may even lead to new discoveries.

% {\appendix[Proof of the Zonklar Equations]
% Use $\backslash${\tt{appendix}} if you have a single appendix:
% Do not use $\backslash${\tt{section}} anymore after $\backslash${\tt{appendix}}, only $\backslash${\tt{section*}}.
% If you have multiple appendixes use $\backslash${\tt{appendices}} then use $\backslash${\tt{section}} to start each appendix.
% You must declare a $\backslash${\tt{section}} before using any $\backslash${\tt{subsection}} or using $\backslash${\tt{label}} ($\backslash${\tt{appendices}} by itself
%  starts a section numbered zero.)}



% \end{thebibliography}

\newpage

 

% \end{IEEEbiography}



\vfill
\newpage

\section{Supplementary}

From the multi-channel EEG data contained in the publicly available CHB-MIT dataset, we extracted the same 16 channels from each subject as this was the largest power-of-two value present in all subjects, allowing us to use the same architecture for everyone. We selected to following channels, as together they uniformly sample all areas of the skull:

\begin{enumerate}
\item FP1-F7
\item FP1-F3
\item FP2-F4
\item FP2-F8
\item F7-T7
\item F3-C3
\item F4-C4
\item F8-T8
\item T7-P7
\item C3-P3
\item C4-P4
\item T8-P8
\item P7-O1
\item P3-O1
\item P4-O2
\item P8-O2
\end{enumerate}

\clearpage

\begin{table}[h!]
  \caption{Performance : Results of Hyperparameter sweep using CNN-Dense on BIHMIT dataset}
  \label{tab:metricsEyetracking}
  \centering
  \begin{tabular}{| l l l l l l r r r |}
  \toprule
  Conv1&Conv2&MPool&nCh1&nCh2&hSize&Params&MACS&Accuracy \\
  \midrule
1x8&2x8&-&1$\rightarrow$8&8$\rightarrow$64&2&137,616&8,533,380&90\% \\
1x8&2x8&-&1$\rightarrow$8&8$\rightarrow$64&5&331,545&8,727,306&91\% \\
1x8&2x8&-&1$\rightarrow$8&8$\rightarrow$64&10&654,760&9,050,516&91\% \\
1x8&2x8&-&1$\rightarrow$8&8$\rightarrow$64&20&1,301,190&9,696,936&92\% \\
1x8&2x8&-&1$\rightarrow$8&8$\rightarrow$64&50&3,240,480&11,636,196&92\% \\
1x8&2x8&-&1$\rightarrow$8&8$\rightarrow$64&100&6,472,630&14,868,296&90\% \\
&&&&&&&& \\
1x1&2x1&-&1$\rightarrow$8&8$\rightarrow$64&-&132,178&1,196,032&71\% \\
1x2&2x2&-&1$\rightarrow$8&8$\rightarrow$64&-&132,954&2,256,608&70\% \\
1x4&2x4&-&1$\rightarrow$8&8$\rightarrow$64&-&134,506&4,365,376&84\% \\
1x8&2x8&-&1$\rightarrow$8&8$\rightarrow$64&-&137,610&8,533,376&92\% \\
1x16&2x16&-&1$\rightarrow$8&8$\rightarrow$64&-&143,818&16,671,232&95\% \\
&&&&&&&& \\
2x1&-&-&1$\rightarrow$8&-&20&163,926&180,264&85\% \\
2x2&-&-&1$\rightarrow$8&-&20&163,782&196,456&85\% \\
2x4&-&-&1$\rightarrow$8&-&20&163,494&228,744&86\% \\
2x8&-&-&1$\rightarrow$8&-&20&162,918&292,936&87\% \\
2x16&-&-&1$\rightarrow$8&-&20&161,766&419,784&88\% \\
&&&&&&&& \\
2x8&-&-&1$\rightarrow$1&-&20&20,419&36,652&77\% \\
2x8&-&-&1$\rightarrow$2&-&20&40,776&73,264&82\% \\
2x8&-&-&1$\rightarrow$4&-&20&81,490&146,488&85\% \\
2x8&-&-&1$\rightarrow$8&-&20&162,918&292,936&87\% \\
2x8&-&-&1$\rightarrow$16&-&20&325,774&585,832&86\% \\
&&&&&&&& \\
1x8&-&2x8&1$\rightarrow$8&-&2&16,240&146,340&71\% \\
1x8&-&2x8&1$\rightarrow$8&-&5&40,489&170,586&77\% \\
1x8&-&2x8&1$\rightarrow$8&-&10&80,904&210,996&82\% \\
1x8&-&2x8&1$\rightarrow$8&-&20&161,734&291,816&86\% \\
1x8&-&2x8&1$\rightarrow$8&-&50&404,224&534,276&87\% \\
1x8&-&2x8&1$\rightarrow$8&-&100&808,374&938,376&88\% \\
&&&&&&&& \\
1x1&-&2x1&1$\rightarrow$8&-&-&16,402&32,768&67\% \\
1x2&-&2x2&1$\rightarrow$8&-&-&16,378&49,088&68\% \\
1x4&-&2x4&1$\rightarrow$8&-&-&16,330&81,632&68\% \\
1x8&-&2x8&1$\rightarrow$8&-&-&16,234&146,336&68\% \\
1x16&-&2x16&1$\rightarrow$8&-&-&16,042&274,208&73\% \\
\bottomrule
  \end{tabular}
\end{table}

\clearpage

\begin{table}[h!]
  \caption{Performance : Results of Hyperparameter sweep using CNN-LSTM on BIHMIT dataset}
  \label{tab:metricsEyetracking}
  \centering
  \begin{tabular}{| l l l l l l l r r r |}
  \toprule
  Conv1&Conv2&MPool&nCh1&nCh2&stride&LSTM-hSize&Params&MACS&Accuracy \\
  \midrule
2x8&-&-&1$\rightarrow$8&-&1&1&230&235,948&83\% \\
2x8&-&-&1$\rightarrow$8&-&1&2&338&357,992&83\% \\
2x8&-&-&1$\rightarrow$8&-&1&5&758&821,756&87\% \\
2x8&-&-&1$\rightarrow$8&-&1&10&1,778&1,920,136&86\% \\
2x8&-&-&1$\rightarrow$8&-&1&20&5,018&5,337,296&86\% \\
2x8&-&-&1$\rightarrow$8&-&1&50&24,338&25,351,976&86\% \\
2x8&-&-&1$\rightarrow$8&-&1&100&88,538&91,253,776&86\% \\
&&&&&&&&& \\
2x8&-&-&1$\rightarrow$1&-&1&2&107&130,184&63\% \\
2x8&-&-&1$\rightarrow$2&-&1&2&140&162,728&65\% \\
2x8&-&-&1$\rightarrow$4&-&1&2&206&227,816&75\% \\
2x8&-&-&1$\rightarrow$8&-&1&2&338&357,992&85\% \\
2x8&-&-&1$\rightarrow$12&-&1&2&470&488,168&86\% \\
2x8&-&-&1$\rightarrow$16&-&1&2&602&618,344&87\% \\
&&&&&&&&& \\
2x8&-&-&1$\rightarrow$8&-&1&2&338&357,992&86\% \\
2x8&-&-&1$\rightarrow$8&-&2&2&338&179,176&86\% \\
2x8&-&-&1$\rightarrow$8&-&4&2&338&89,768&86\% \\
2x8&-&-&1$\rightarrow$8&-&8&2&338&45,064&84\% \\
2x8&-&-&1$\rightarrow$8&-&16&2&338&22,536&78\% \\
2x8&-&-&1$\rightarrow$8&-&32&2&338&11,272&72\% \\
&&&&&&&&& \\
1x8&2x8&-&1$\rightarrow$8&8$\rightarrow$64&1&2&9,426&9,535,304&86\% \\
1x8&2x8&-&1$\rightarrow$8&8$\rightarrow$64&2&2&9,426&2,402,472&91\% \\
1x8&2x8&-&1$\rightarrow$8&8$\rightarrow$64&4&2&9,426&609,992&87\% \\
1x8&2x8&-&1$\rightarrow$8&8$\rightarrow$64&8&2&9,426&165,384&85\% \\
1x8&2x8&-&1$\rightarrow$8&8$\rightarrow$64&12&2&9,426&76,072&78\% \\
1x8&2x8&-&1$\rightarrow$8&8$\rightarrow$64&16&2&9,426&45,448&69\% \\
&&&&&&&&& \\
1x8&2x8&-&1$\rightarrow$1&1$\rightarrow$1&1&2&116&145,560&57\% \\
1x8&2x8&-&1$\rightarrow$2&2$\rightarrow$4&1&2&288&323,432&78\% \\
1x8&2x8&-&1$\rightarrow$4&4$\rightarrow$16&1&2&1,406&1,454,856&88\% \\
1x8&2x8&-&1$\rightarrow$8&8$\rightarrow$64&1&2&9,426&9,535,304&83\% \\
&&&&&&&&& \\
1x8&2x8&-&1$\rightarrow$12&12$\rightarrow$144&1&2&30,278&30,543,752&70\% \\
1x8&2x8&-&1$\rightarrow$16&16$\rightarrow$256&1&2&70,106&70,685,640&67\% \\
1x8&2x8&-&1$\rightarrow$20&20$\rightarrow$400&1&2&135,054&136,166,408&64\% \\
&&&&&&&&& \\
1x8&-&2x8&1$\rightarrow$1&-&1&2&99&129,400&60\% \\
1x8&-&2x8&1$\rightarrow$2&-&1&2&124&161,832&64\% \\
1x8&-&2x8&1$\rightarrow$4&-&1&2&174&226,696&64\% \\
1x8&-&2x8&1$\rightarrow$8&-&1&2&274&356,424&69\% \\
1x8&-&2x8&1$\rightarrow$12&-&1&2&374&486,152&72\% \\
1x8&-&2x8&1$\rightarrow$16&-&1&2&474&615,880&67\% \\
1x8&-&2x8&1$\rightarrow$20&-&1&2&574&745,608&67\% \\
&&&&&&&&& \\
1x8&-&2x8&1$\rightarrow$8&-&1&2&274&356,424&71\% \\
1x8&-&2x8&1$\rightarrow$8&-&2&2&274&121,384&72\% \\
1x8&-&2x8&1$\rightarrow$8&-&4&2&274&46,536&73\% \\
1x8&-&2x8&1$\rightarrow$8&-&8&2&274&19,976&72\% \\
1x8&-&2x8&1$\rightarrow$8&-&12&2&274&12,456&70\% \\
1x8&-&2x8&1$\rightarrow$8&-&16&2&274&9,096&70\% \\
\bottomrule
  
  \end{tabular}
\end{table}

\clearpage

\begin{table}[h!]
  \caption{Performance : Results of Hyperparameter sweep using Scattering Network on BIHMIT dataset}
  \label{tab:metricsEyetracking}
  \centering
  \begin{tabular}{| l l l r r r |}
  \toprule
  j&q&hSize&Params&MACs&Accuracy \\
  \midrule
1&8&10&45,088&81,940&69\% \\
2&8&10&24,608&67,604&72\% \\
4&8&10&36,128&152,340&85\% \\
6&8&10&24,288&221,68&89\% \\
8&8&10&13,488&292,00&91\% \\
&&&&& \\
4&8&2&10,504&55,556&66\% \\
4&8&5&20,113&65,162&76\% \\
4&8&10&36,128&81,172&86\% \\
4&8&20&68,158&113,192&89\% \\
4&8&50&164,248&209,252&92\% \\
4&8&100&324,398&369,352&93\% \\
&&&&& \\
4&1&10&12,448&59,028&80\% \\
4&2&10&16,928&79,892&83\% \\
4&4&10&22,688&106,132&84\% \\
4&6&10&29,728&133,652&84\% \\
4&8&10&36,128&152,340&85\% \\
&&&&& \\
4&8&2&10,504&55,556&74\% \\
4&8&5&20,113&65,162&76\% \\
4&8&10&36,128&81,172&88\% \\
4&8&20&68,158&113,192&92\% \\
4&8&50&164,248&209,252&94\% \\
4&8&100&324,398&369,352&94\% \\
&&&&& \\
1&8&10&45,088&81,940&70\% \\
2&8&10&24,608&67,604&72\% \\
4&8&10&36,128&152,340&88\% \\
6&8&10&24,288&221,268&93\% \\
8&8&10&13,488&292,100&95\% \\
&&&&& \\
4&1&10&12,448&59,028&86\% \\
4&2&10&16,928&79,892&89\% \\
4&4&10&22,688&106,132&88\% \\
4&6&10&29,728&133,652&89\% \\
4&8&10&36,128&152,340&88\% \\
\bottomrule
  
  \end{tabular}
\end{table}

\clearpage

\begin{table}[h!]
  \caption{Performance : Results of Hyperparameter sweep using Time Scale Network on BIHMIT dataset}
  \label{tab:metricsEyetracking}
  \centering
  \begin{tabular}{| l l l r r r |}
  \toprule
  minWin&maxWin&numCh&Params&MACS&Accuracy \\
  \midrule
4&2048&1&13,327&30,709&79\% \\
4&1024&1&9,230&27,638&78\% \\
4&512&1&7,181&24,567&77\% \\
4&256&1&6,156&21,496&75\% \\
4&128&1&5,643&18,425&73\% \\
&&&&& \\
4&2048&1&13,327&30,709&79\% \\
8&2048&1&10,762&23,030&79\% \\
16&2048&1&9,473&18,423&78\% \\
32&2048&1&8,816&15,224&78\% \\
64&2048&1&8,463&12,665&76\% \\
128&2048&1&8,238&10,394&75\% \\
&&&&& \\
4&2048&1&13,327&30,709&79\% \\
8&1024&1&6,665&20,471&78\% \\
16&512&1&3,327&13,817&76\% \\
32&512&1&2,670&10,874&75\% \\
&&&&& \\
4&2048&2&26,652&61,418&82\% \\
8&1024&2&13,328&40,942&82\% \\
16&512&2&6,652&27,634&80\% \\
32&512&2&5,338&21,748&80\% \\
64&512&2&4,632&16,886&78\% \\
64&256&2&2,582&12,664&76\% \\
&&&&& \\
4&2048&1&13,327&30,709&79\% \\
4&2048&3&39,977&92,127&85\% \\
4&2048&5&66,627&153,545&85\% \\
4&2048&7&93,277&214,963&86\% \\
4&2048&10&133,252&307,090&85\% \\
&&&&& \\
4&2048&3&39,977&92,127&84\% \\
4&1024&3&27,686&82,914&84\% \\
4&512&3&21,539&73,701&81\% \\
4&256&3&18,464&64,488&80\% \\
4&128&3&16,925&55,275&78\% \\
&&&&& \\
4&2048&3&39,977&92,127&84\% \\
8&2048&3&32,282&69,090&85\% \\
16&2048&3&28,415&55,269&85\% \\
32&2048&3&26,444&45,672&84\% \\
64&2048&3&25,385&37,995&84\% \\
128&2048&3&24,710&31,182&82\% \\
&&&&& \\
4&2048&3&39,977&92,127&84\% \\
8&1024&3&19,991&61,413&84\% \\
16&512&3&9,977&41,451&82\% \\
32&512&3&8,006&32,622&82\% \\
64&512&3&6,947&25,329&81\% \\
128&512&3&6,272&18,708&79\% \\
256&512&3&5,645&12,375&77\% \\
128&256&3&3,197&12,471&76\% \\
&&&&& \\
4&2048&5&66,627&153,545&85\% \\
4&1024&5&46,142&138,190&85\% \\
4&512&5&35,897&122,835&83\% \\
4&256&5&30,772&107,480&81\% \\
4&128&5&28,207&92,125&78\% \\
&&&&& \\
4&2048&5&66,627&153,545&86\% \\
8&2048&5&53,802&115,150&86\% \\
16&2048&5&47,357&92,115&86\% \\
32&2048&5&44,072&76,120&87\% \\
64&2048&5&42,307&63,325&86\% \\
128&2048&5&41,182&51,970&85\% \\
\bottomrule

  \end{tabular}
\end{table}

\clearpage

\begin{table}[h!]
  \caption{Performance : Results of Hyperparameter sweep using Time Scale Network on CHBMIT dataset}
  \label{tab:metricsEyetracking}
  \centering
  \begin{tabular}{| l l l r r r |}
  \toprule
  minWin&maxWin&numCh&Params&MACS&Accuracy \\
  \midrule
32&16384&1&35,862&181,237&90.1\% \\
32&4096&1&11,284&146,423&91.0\% \\
32&1024&1&5,138&111,609&91.9\% \\
32&256&1&3,600&76,795&91.0\% \\
&&&&& \\
32&16384&1&35,862&174,069&90.3\% \\
64&16384&1&34,325&152,054&90.2\% \\
128&16384&1&33,556&133,111&89.7\% \\
256&16384&1&33,171&115,576&89.1\% \\
512&16384&1&32,978&98,681&87.2\% \\
1024&16384&1&32,881&82,074&84.0\% \\
&&&&& \\
128&512&1&1,807&49,916&91.8\% \\
256&512&1&\bf{1,422}&33,021&90.9\% \\
128&1024&1&2,832&66,555&91.8\% \\
256&1024&1&2,447&49,532&91.3\% \\
512&1024&1&2,254&32,893&88.7\% \\
&&&&& \\
32&1024&1&5,138&104,441&91.81\% \\
32&1024&2&10,274&209,904&\bf{92.26}\% \\
32&1024&3&15,410&315,367&92.25\% \\
32&1024&5&25,682&524,249&92.22\% \\
\bottomrule

  \end{tabular}
\end{table}

\clearpage

\begin{table}[h!]
  \caption{Performance : Subject-specific results for the best Time Scale Network (3 channels each containing a TiSc input layer with $\Lambda=[5,10]$, TiSc hidden layer with $\Lambda=[6,10]$, densely connected to a one-hot output) and the Time Scale Network with the fewest parameters (1 channel containing a TiSc input layer with $\Lambda=[8,9]$, hidden layer with $\Lambda=[9]$, densely connected to a one-hot output) on CHBMIT dataset}
  \label{tab:metricsEyetracking}
  \centering
  \begin{tabular}{| l r r |}
  \toprule
  Subject&Best Accuracy&Minimum Parameters \\
  \midrule
chb01&99.4\%&99.1\% \\
chb02&95.1\%&92.3\% \\
chb03&91.3\%&87.1\% \\
chb04&86.9\%&77.9\% \\
chb05&92.0\%&86.0\% \\
chb06&90.8\%&90.7\% \\
chb07&94.4\%&91.6\% \\
chb09&93.4\%&90.6\% \\
chb10&95.1\%&93.7\% \\
chb11&93.1\%&91.8\% \\
chb13&97.3\%&96.7\% \\
chb14&95.1\%&95.1\% \\
chb15&92.3\%&95.2\% \\
chb16&92.3\%&87.4\% \\
chb17&94.3\%&91.3\% \\
chb18&88.0\%&85.7\% \\
chb19&97.7\%&97.5\% \\
chb20&97.9\%&96.1\% \\
chb21&83.9\%&81.4\% \\
chb22&79.5\%&85.0\% \\
chb23&97.7\%&96.5\% \\
\bottomrule

  \end{tabular}
\end{table}

\end{document}